\documentclass[english,conference]{ieeeconf}

\usepackage[english]{babel}
\usepackage[utf8]{inputenc}
\usepackage[hidelinks]{hyperref}
\usepackage{amsmath,amsfonts}
\usepackage[font=footnotesize]{subcaption}
\usepackage[font=small,labelfont=bf]{caption}
\usepackage{cleveref}
\usepackage{algpseudocode}
\usepackage{tikz}
\usepackage{bm}
\usepackage{algorithm}
\usepackage{bbm}
\usepackage{nicefrac}
\usepackage[absolute]{textpos}
\usepackage[prependcaption=inline]{todonotes}
\usepackage{microtype}

\usetikzlibrary{
	arrows,
	arrows.meta,
	angles,
	babel,
	backgrounds,
	calc,
	decorations,
	decorations.pathreplacing,
	decorations.pathmorphing,
	fit,
	intersections,
	matrix,
	petri,
	positioning,
	shapes,
	spy,
	through,
	quotes,
}

\newcommand{\copyrightstatement}{
	\begin{textblock*}{17cm}(20mm,1mm)    
		\noindent
		\footnotesize
		\copyright  Copyright 2018 IEEE. Published in the IEEE 2018 Global Conference on Signal and Information Processing (GlobalSIP 2018), scheduled for November 26-28, 2018 in Anaheim, California, USA. Personal use of this material is permitted. However, permission to reprint/republish this material for advertising or promotional purposes or for creating new collective works for resale or redistribution to servers or lists, or to reuse any copyrighted component of this work in other works, must be obtained from the IEEE. Contact: Manager, Copyrights and Permissions / IEEE Service Center / 445 Hoes Lane / P.O. Box 1331 / Piscataway, NJ 08855-1331, USA. Telephone: + Intl. 908-562-3966. DOI: 10.1109/GlobalSIP.2018.8646554
	\end{textblock*}
}

\RequirePackage{pgfplots,pgfplotstable}
\pgfplotsset{compat=newest}
\usepgfplotslibrary{units,groupplots,colormaps,fillbetween}
\DeclareMathOperator*{\argmax}{arg\,max}

\definecolor{tasblue}{RGB}{25,171,255}

\IEEEoverridecommandlockouts

\begin{document}
\copyrightstatement
\bstctlcite{bibcontrol_etal4}
\title{%
    The Greedy Dirichlet Process Filter - \\An Online Clustering Multi-Target Tracker
}

\author{
    Benjamin Naujoks%
    \thanks{
        All authors are with the Institute for Autonomous Systems Technology (TAS) of the Universität der Bundeswehr Munich, Neubiberg, Germany. Contact author email: benjamin.naujoks@unibw.de},
    Patrick Burger
    and %
    Hans-Joachim Wuensche
}

\maketitle
\begin{abstract}
    Reliable collision avoidance is one of the main requirements for autonomous driving.
Hence, it is important to correctly estimate the states of an unknown number of static and dynamic objects in real-time.
Here, data association is a major challenge for every multi-target tracker.
We propose a novel multi-target tracker called Greedy Dirichlet Process Filter (GDPF) based on the non-parametric Bayesian model called Dirichlet Processes and the fast posterior computation algorithm Sequential Updating and Greedy Search (SUGS).
By adding a temporal dependence we get a real-time capable tracking framework without the need of a previous clustering or data association step.
Real-world tests show that GDPF outperforms other multi-target tracker in terms of accuracy and stability.\\

\end{abstract}
\section{Introduction}
Every autonomous car needs a collision-avoidance system.
Therefore, it has to detect and track static and dynamic objects.
In this context, single-target tracking is well researched and a lot of progress has been made in the past decades \cite{bena:BarShalom02,bena:mahlerstatistics,BayesianFiltering}.

However, multi-target tracking is a far more challenging task, as there is an unknown and time-changing number of targets \cite{bena:mahlerstatistics,bena:LMB}.
In every time step, new targets have to be initialized or pruned.
Additionally, there exist clutter measurements which are not associated to any target.
Furthermore, targets are temporally occluded, due to the specific sensor field of view.
Consequently, the measurement to target association is a major challenge and multi-target tracker have to tackle all the previously mentioned aspects.

\begin{figure}[t!]
	\begin{center}
		\includegraphics[width=0.45\textwidth]{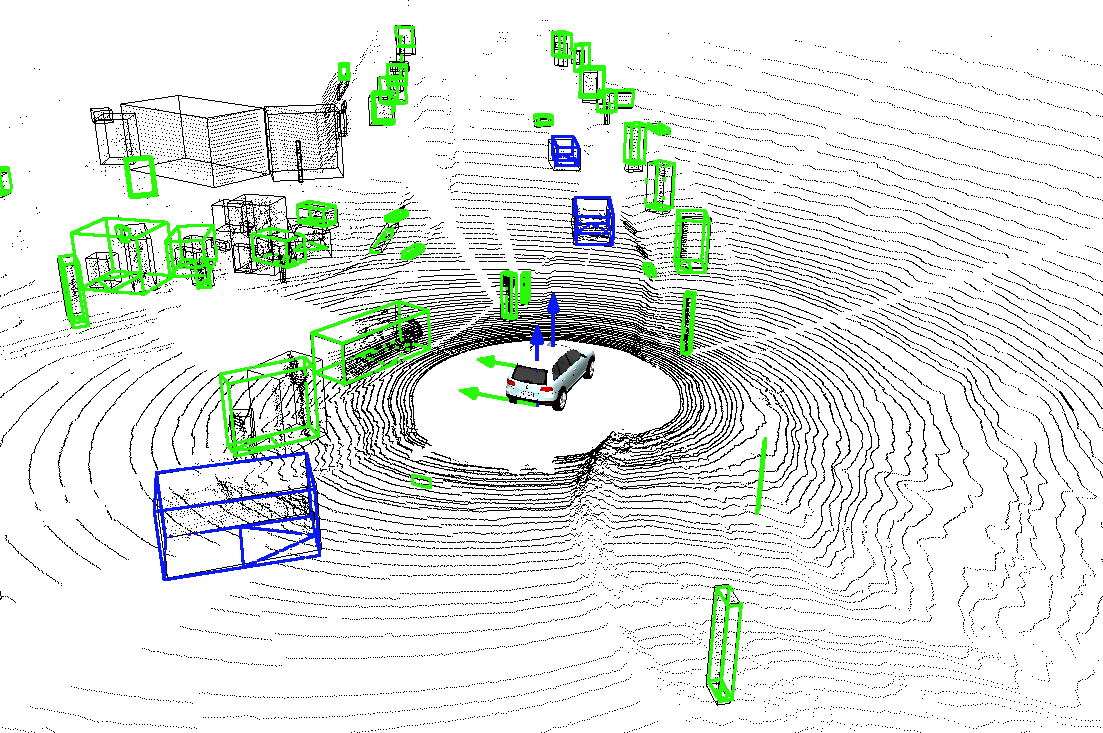}
	\end{center}
	\caption{The result of the GDPF in a suburban area. Green boxes denote alive static targets, whereas blue boxes are moving targets.}\label{fig:front}
	\vspace{-0.3cm}
\end{figure}

In this paper, we propose a novel filter framework, GDPF, based on non-parametric Bayesian model called Dirichlet Processes \cite{bena:bleidistance} and the fast posterior computation algorithm Sequential Updating and Greedy Search (SUGS) \cite{Wang_fastbayesian}.
The GDPF handles the data association in a probabilistic manner without the need of a previous clustering step.
Furthermore, GDPF can deal with over segmentation of a previous clustering step.
\Cref{fig:front} shows an exemplary use case of the GDPF in a suburban area, which was recorded with a roof-mounted Velodyne HDL64-S2 \cite{velodyne-lidar} of our institute's autonomous car MuCAR-3 \cite{bena:fries2017itsc}.

This paper is structured as follows:
\Cref{sec:related_works} starts with discussing related work.
Next, \Cref{sec:filter} shows the GDPF algorithm and explains its single steps.
Moreover, in \Cref{sec:results} the performance of the proposed filter is evaluated against other filter frameworks in a real-world scenario.
Lastly, \Cref{sec:conclusion} summarizes the paper and proposes future work.

\section{Related Work} \label{sec:related_works}
Multi-target tracker can be divided into different measurement processing strategies.
One strategy for LiDAR sensors is to firstly segment the point cloud into ground plane as well as obstacles and then cluster the remaining obstacle points into coherent objects \cite{bena:burger_iv2018,bena:BeNaIV2018}.
Afterwards, these objects are used as measurements in multi-target tracking \cite{bena:LMB,vo2006gaussian,bena:vocardinality}.
The disadvantages of this approach are the need of an additional clustering algorithm and the lack of implicit handling of possible segmentation errors.

Another strategy is called extended target tracking.
Here, the target's appearance is modeled in the tracking algorithm and multiple detections can be associated to targets.
Multi-Hypothesis tracker can be extended for this use case \cite{bena:LMB,granstromextended,bena:lmbextended}.
Our main contribution is a multi-target tracker with probabilistic association approach which utilizes the knowledge of previous associations through the usage of Dirichlet Processes.
Furthermore, it is applicable with both mentioned measurement processing strategies.
In contrast to other DP based methods, which utilize Monte-Carlo based methods \cite{bena:tuncersequential,bena:foxddsdlm}, we use the fast greedy posterior inference of the clustering method SUGS \cite{Wang_fastbayesian} and extend the algorithm with a temporal dependence.
Lastly, the GDPF works under real-time constraints even for hundreds of targets.
\section{Filter} \label{sec:filter}
\subsection{Temporally-Dependent Dirichlet Process Mixture Model}
Before our GDPF algorithm is introduced some basic definitions are explained.
A Dirichlet Process (DP) is a Bayesian non-parametric model.
Basically, it is a distribution over distributions.
Furthermore, DPs have an infinite amount of mixture components, but only finite ones are activated by observations.
\subsubsection{Chinese Restaurant Process}
The Chinese Restaurant Process (CRP) is a realization of a DP.
Figuratively speaking, the CRP can be interpreted as follows:
A new customer (measurement) either chooses a new table with a probability proportional to concentration parameter $ \alpha \in\mathbb{R} $ or joins a previously known table with a probability proportional to the number of occupying customers at the table \cite{Wang_fastbayesian}.
Formalizing this and by marginalizing out the random mixing measure $ G $, we obtain the DP prediction rule for the assignments of the measurements $ \bm{y}_i(t) $ with $ i=1,\dots,n $, $ n\in \mathbb{N} $ is the number of measurements and time step $ t $ \cite{Wang_fastbayesian}:
\begin{multline}
\bm{\theta}_{z_i(t)}(t)|\bm{\theta}_{z_0(t)}(t),\dots,\bm{\theta}_{z_{i-1}(t)}(t) \sim  \\
 \sum\limits_{k \in K_t} \frac{n_k(t)}{i-1 + \alpha} \delta(\bm{\theta}_{k} (t))+\frac{\alpha}{i-1+\alpha}G_0, \label{eq:dp_predict}
\end{multline}
where $ z_i(t): \mathbb{N} \to K_t $ is the cluster indicator, which denotes the assignment of measurement $ \bm{y}_i $ to cluster index $ k $, $ K_t $ is the set of cluster indices, $\bm{\theta}_k $ are cluster parameters specific to the cluster with index $ k  \in K_t$, $ \delta(\bm{\theta}_{k}(t)) $ is the delta distribution around $ \bm{\theta}_{k}(t) $, $ n_k(t) $ is the number of assigned measurements to the cluster with index $ k $ and $ G_0 $ is the DP base prior.
Now, we can define with \Cref{eq:dp_predict} the conditional prior distribution on the cluster assignments for sequentially assigning measurements to the clusters \cite{Wang_fastbayesian}:
\begin{equation}
CRP(\alpha) = \begin{cases}
	 \frac{n_k(t)}{ i-1+\alpha} &k \in K_t,\\
	 \frac{\alpha}{ i-1+\alpha } &\text{else}
 \end{cases}.
\end{equation}
\subsubsection{Distance Dependent Chinese Restaurant Process}
The distance dependent Chinese Restaurant Process (ddCRP) is a direct extension of the CRP. It can describe a distribution over partitions indirectly via distributions over links between different data measurements. Figuratively speaking, the ddCRP links customer to other customers rather than tables \cite{bena:bleidistance,bena:ddmotionseg}.
It follows the ddCRP version of the conditional prior distribution for time step $ t $:
\begin{equation}
p( j_i(t) = l | \bm{j}_{-i},\alpha ) = \begin{cases}
d_{il}(\bm{y}_i (t),\bm{y}_l (t) )  &i \neq l, \\
\alpha  &i=l
\end{cases}, \label{eq:ddcrp_prior}
\end{equation} where  $ j_i(t) $ is the link assignment between two measurements $ \bm{y}_i $, $ \bm{y}_l $, all previous link assignments are denoted by $ \bm{j}_{-i} $ and $ d_{il}(\bm{y}_i(t),\bm{y}_l(t) ) $ is a distance dependent probability function.
\subsubsection{Data To Cluster Assignments}
The data to cluster assignment is dependent on the current cluster parameters and previous assignments.
Therefore, the likelihood is defined as follows:
\begin{equation}
	\pi_{\bm{z}(t)} = \lbrace \pi_{ k} \rbrace_{k\in K_t},
\end{equation} where $ \bm{z}(t) $ denotes all measurement to cluster assignments for time step $ t $.
Moreover, the transition between different time steps depending on measurement $ \bm{y}_i $ is defined as:
\begin{align}
\pi_{z_i(t)=k} &= p\left(z_i(t)=k\vert \bm{z}(t-1),\bm{y}_i(t)\right) \nonumber \\
				 &= p( \bm{y}_i(t)\vert \bm{\theta}_k(t) )\cdot p\left(z_i(t)=k\vert \bm{z}(t-1)\right), \label{eq:data_assi}
\end{align} where $p\left(\bm{y}_i(t)\vert \bm{\theta}_k(t)\right)$ is the likelihood of measurement $ \bm{y}_i(t) $ dependent on cluster parameters $ \theta_k(t) $ and $p\left(z_i(t)=k\vert \bm{z}(t-1)\right)$ is the transition of the cluster assignments between two time steps, which mostly is defined as uniformly distributed.
\subsubsection{Kalman Filtering}
For state estimation of the dynamical part $ \bm{x}\in\mathbb{R}^n $ of the mixture components, we use the standard Kalman Filtering approach.
Therefore, consider the following probabilistic and Gaussian state space model for measurement $ \bm{y} $ and time step $ t $ \cite{bena:MU}:
\begin{align}
p(\bm{x}(t)\vert \bm{x}(t-1)) &= \mathcal{N}\left(\bm{x}(t)\vert \Phi(t-1)\bm{x}(t-1),\mathbf{Q}(t-1)\right)\\
p(\bm{y}(t)\vert \bm{x}(t)) &= \mathcal{N}\left( \bm{y}(t)\vert \mathbf{C}(t) \bm{x}(t),\mathbf{R}(t)\right),
\end{align} where $ \Phi(t-1) \in \mathbb{R}^{n\times n}$ is the transition matrix, $ \mathbf{C}(t)\in \mathbb{R}^{m \times n} $ is the measurement model matrix, $ \mathbf{Q}(t-1) $ is the unbiased and Gaussian process noise and $ \mathbf{R}(t) $ is the unbiased and Gaussian measurement noise.
Naturally, this part can be extended for non-linear Kalman Filtering.
For example, $\Phi(t-1)$ and $\mathbf{C}(t)$ of non-linear dynamic, respectively measurement models could be approximated by Extended Kalman Filtering techniques.
\subsection{Greedy Dirichlet Process Filter}
\begin{figure}[t!]
	\vspace{-1cm}
	\begin{center}
		\begin{tikzpicture}[scale=0.8, transform shape]
\tikzstyle{state}=[shape=circle,draw=blue!50,fill=blue!20]
\tikzstyle{repetition}=[shape=circle,draw=white!100]
\tikzstyle{lightedge}=[<-,dotted]
\tikzstyle{mainstate}=[state,thick]
\tikzstyle{mainedge}=[<-,thick]
\tikzset{
	main/.style={circle, minimum size = 4mm, thick, draw =black!80, node distance = 10mm},
	connect/.style={-latex, thick},
	box/.style={rectangle, draw=black!100}
}
\node[repetition](fix){};
\node[repetition](fix15)[below=of fix]{};
\node[repetition](fix2)[below=of fix15]{};
\node[main] (pi) at ($(fix)!0.5!(fix2)$) { $ \pi_k $ };
\node[draw=black,inner sep=2mm,thick,rectangle,fit=(pi)] (pibox) {};
\node[anchor=south east,inner sep=1pt] at (pibox.south east)
{$ \infty$};
\node[main] (Z1) [right=of fix2] {$\bm{z}_0$};
\node[main] (Z2) [right=of Z1] {$\bm{z}_1$};
\node[main] (Zt) [right=of Z2] {$\bm{z}_t$};
\node[main,below=of pi ] (Theta) {$ \theta_k $};
\node[draw=black,inner sep=2mm,thick,rectangle,fit=(Theta)] (Thetabox) {};
\node[anchor=south east,inner sep=1pt] at (Thetabox.south east)
{$ \infty$};
\node[main] (L2) [below=of Z1] {$\bm{x}_0$};
\node[main] (L3) [right=of L2] {$\bm{x}_1$};
\node[main] (Lt) [right=of L3] {$\bm{x}_t$};
\node (G0) [left=of Theta]{$ G_0 $};
\node (alpha)[left=of pi]{$ \alpha $};

\node[main,fill=black!10] (O2) [below=of L2] {$\bm{y}_0$};
\node[main,fill=black!10] (O3) [below=of L3] {$\bm{y}_1$};
\node[main,fill=black!10] (Ot) [below=of Lt] {$\bm{y}_t$};
\node[box,draw=white!100,left=of O2] (Observed) {\textbf{Observations}};
\path (Z1) edge [connect] (Z2);
\path (Z2) -- node[auto=false]{\ldots} (Zt);
\path (L3) -- node[auto=false]{\ldots} (Lt);
\path (L2) edge [connect] (L3)
(L3) -- node[auto=false]{\ldots} (Lt);
(O2) edge [connect] (O3)
(O3) -- node[auto=false]{\ldots} (Ot);
\path (L2) edge [connect] (O2);
\path (L3) edge [connect] (O3);
\path (Lt) edge [connect] (Ot);
\path (Z1) edge [connect] (L2);
\path (Z2) edge [connect] (L3);
\path (Zt) edge [connect] (Lt);

\path (pi) edge [connect] (Z1.north);
\path (pi) edge [connect] (Z2.north);
\path (pi) edge [connect] (Zt.north);
\path (Theta) edge [connect] (L2.north);
\path (Theta) edge [connect] (L3.north);
\path (Theta) edge [connect] (Lt.north);
\path (alpha) edge [connect] (pi);
\path (G0) edge [connect] (Theta);
\draw [dashed, shorten >=-1cm, shorten <=-1cm]
($(Theta)!0.65!(Observed)$) coordinate (a) -- ($(Lt)!(a)!(Ot)$);
\end{tikzpicture}
	\end{center}
	\caption{Graphical model of our GDPF. The numerical indices denote the corresponding time step. Moreover, rectangles denote repetition in the model for every cluster/component. Additionally, $ \alpha $ is the concentration parameter which controls the birth of new clusters and $ G_0 $ is the base distribution of the model. } \label{fig:tddpmm}
	\vspace{-0.5cm}
\end{figure}
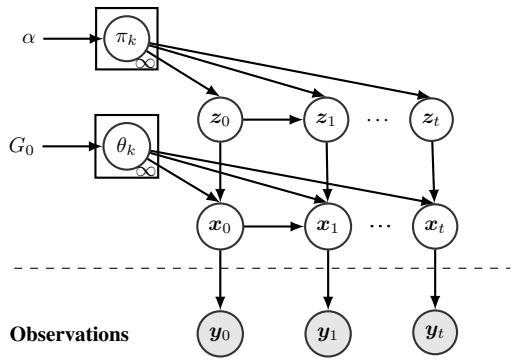

Given the previous definitions, we can finally define our GDPF.
Our proposed GDPF is a direct extension of the SUGS algorithm of \cite{Wang_fastbayesian} by incorporating dynamic and temporal modeling.
The basic procedure of the GDPF is shown in \Cref{alg:gdpf}.
\begin{algorithm}[t!]
	\begin{algorithmic}[1]
		\State $ i=0 $
		\For {$i<\#measurements$}
		\State {Choose best label for measurement $\bm{y}_i$ with:
			\begin{equation}
			k = \argmax\limits_{\hat{k} \in K_t \cup \lbrace k_{new} \rbrace}\left(  p(z_i(t) = \hat{k}|\bm{y}^{i},\bm{j}_{-i},\bm{z}(t-1)) \right) \nonumber
			\end{equation}
		}
		\State {Update posterior distribution $ p( \theta_{z_i(t)}\vert  \bm{y}_{i-1}(t), \bm{z(t)} ) $}
		\EndFor
		\State $pruneComponents(\gamma)$
	\end{algorithmic}
	\caption{GDPF - Greedy Dirichlet Process Filter} \label{alg:gdpf}
\end{algorithm}
Moreover, the underlying model of the filter is illustrated in \Cref{fig:tddpmm}.
In the following the single steps of the filter are defined.
\subsubsection{Choosing the best label}
Let $ \bm{y}_i(t) $ be the i-th measurement for $ i=0,\dots,n $. Then, with  \Cref{eq:ddcrp_prior,eq:data_assi} we get the conditional posterior probability of assigning measurement $ \bm{y}_i $ to cluster $ k $ given the data for measurements $\bm{Y}^{i}(t)=( \bm{y}_0(t),\bm{y}_1(t),\dots,\bm{y}_i(t) ) $ and the previous link assignments $ \bm{j}_{-i} $:
\begin{multline}
 p\left(z_i(t) = k|\bm{Y}^{i}(t),\bm{j}_{-i},\bm{z}(t-1)\right) = \\
 \frac{p(j_i = l | \bm{j}_{-i},\alpha ) \cdot \pi_{z_i(t)=k} }{\sum_{m\in K_t}p(j_i = m | \bm{j}_{-i},\alpha ) \cdot \pi_{z_i(t)=m}}. \label{eq:label}
\end{multline}
\subsubsection{Update}
The cluster parameters are updated according to the following posterior distribution:
\begin{align}
	p( \theta_{z_i(t)}\vert  \bm{y}_{i-1}(t), \bm{z}(t) ) &\propto G_0(\theta_{z_i(t)=k}(t))\\
	&\cdot p( \bm{y}_i(t)\vert \bm{\theta}_{z_i(t)=k}(t)  ) \\
	&\cdot p( \bm{\theta}_{z_i(t)=k}(t )\vert \bm{\theta}_{z_i(t-1)=k}(t-1 )),
\end{align}
where the first part denotes the generation of new components with the base prior distribution, the second part includes the update of the dynamic parameters as well as other cluster parameters, e.g, existence probability. Finally, the third part models the time evolution of the cluster parameters.
\subsubsection{Pruning}
The easiest way to prune components is done by erasing the components which have an existence probability smaller than a death threshold $ \gamma \in\mathbb{R}_{>0}$.

Two exemplary applications are presented in the next section.
\section{Example Applications \& Results} \label{sec:results}
\begin{figure}[t!]
	\begin{center}
		\begin{subfigure}{0.4\textwidth}
			\begin{center}
				\includegraphics[width=0.88\textwidth]{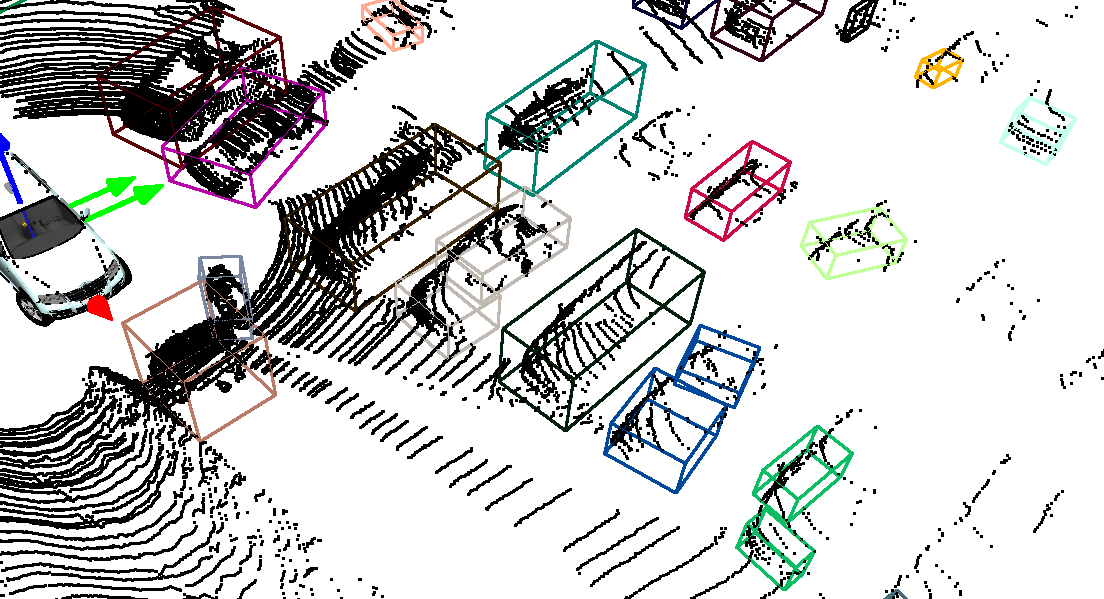}
			
			\caption{Measurement assignments to the corresponding clusters. Different measurements are assigned to the same cluster, which is indicated by color similarity.}\label{fig:bb}
						\end{center}
		    	\vspace{1mm}
		\end{subfigure}
		\begin{subfigure}{0.4\textwidth}
			\begin{center}
				\includegraphics[width=0.88\textwidth]{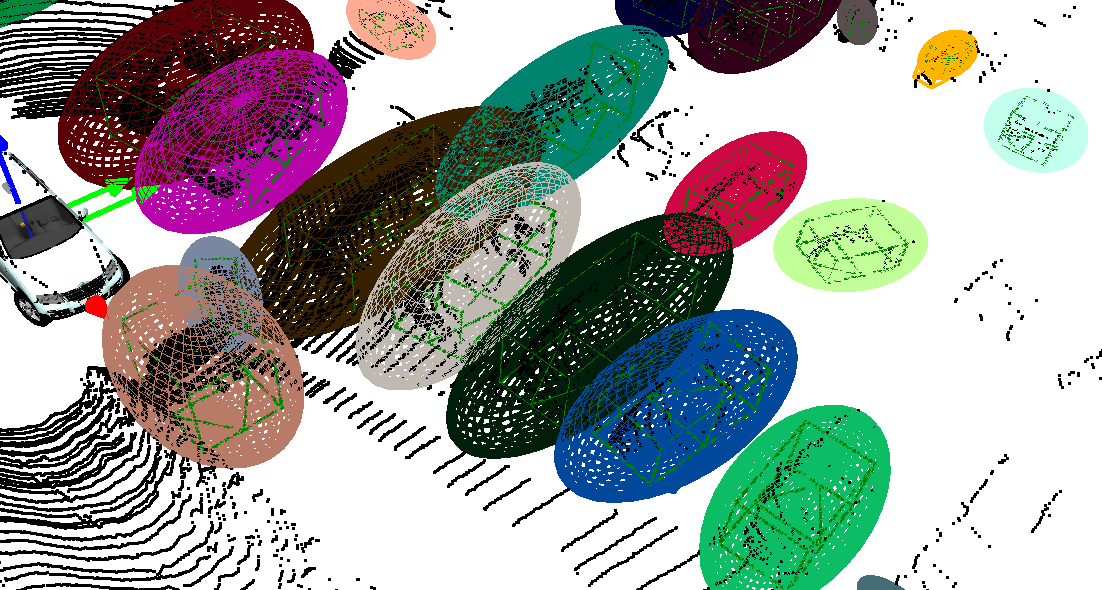}
			\caption{The measurement assignment likelihoods of the bounding boxes depending on the current cluster parameters are illustrated as ellipses. This can be seen as a cluster prior.} \label{fig:bb_ellipse}
				\end{center}
		\end{subfigure}
	\end{center}
	\caption{The GDPF results with clustered bounding boxes as measurements.}
\end{figure}
\begin{figure}[t!]
	\begin{center}
		\begin{subfigure}{0.4\textwidth}
			\begin{center}
				\includegraphics[width=0.88\textwidth]{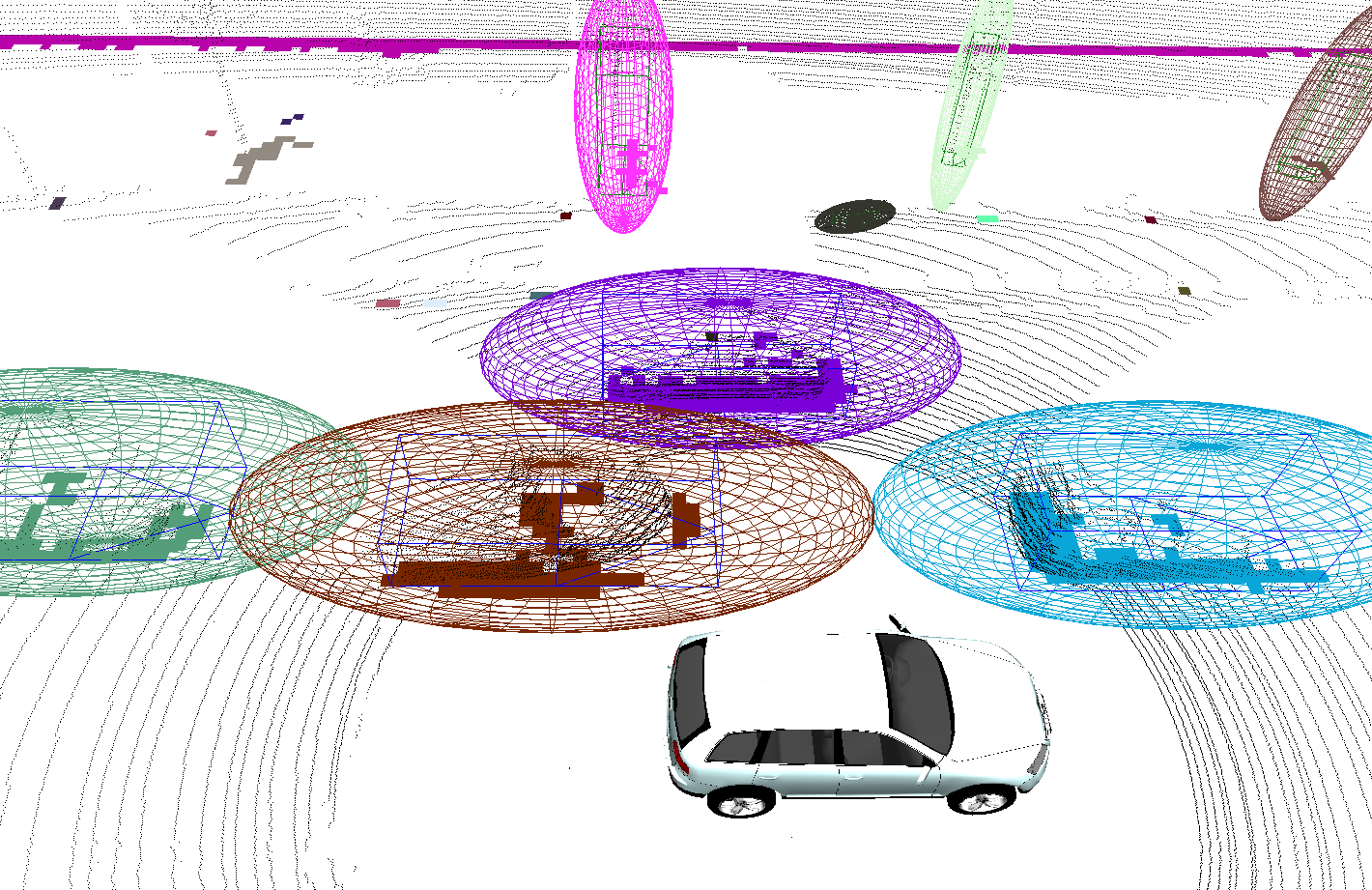}
			\caption{Measurement (grid cells) assignments to the corresponding cluster and their ellipses. The same cluster assignment is denoted by the same color.}\label{fig:grid}
		\end{center}
		\end{subfigure}\\
		\begin{subfigure}{0.4\textwidth}
			\begin{center}
				\includegraphics[width=0.88\textwidth, trim={0 100 0 50},clip]{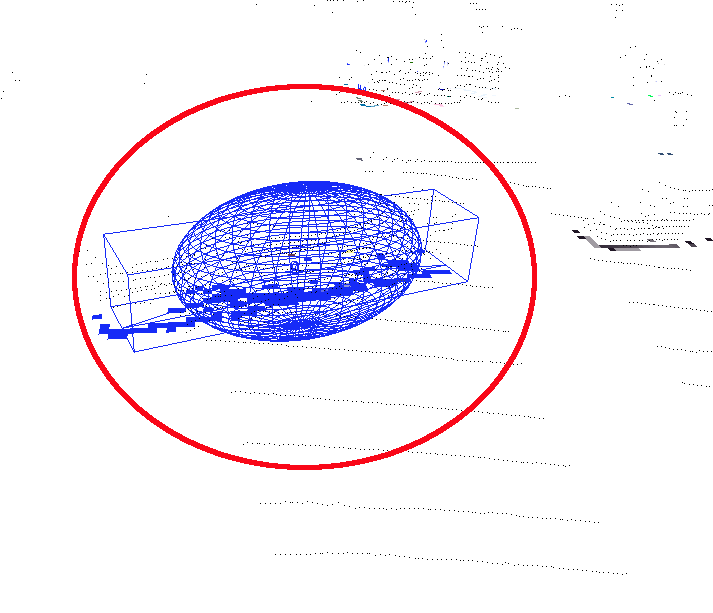}

			\caption{The measurement assignment likelihoods of grid cells depending on the current cluster parameters are illustrated as ellipses, which again can be seen as a cluster prior.}\label{fig:grid_ellipse}
						\end{center}
		\end{subfigure}
	\vspace{0.2cm}
	\caption{The GDPF results with grid cells as measurements. The grid cells in (a) are associated to the correct component through the combination of the prior and ddCRP. Furthermore, the object in the red circle at (b) is correctly segmented through the ddCRP, even though the assumed prior, illustrated as the blue ellipse, is wrong.}
		\end{center}
\end{figure}
Our proposed GDPF is examined in a real-world scenario in a suburban area with passing cars as dynamic objects.
Video footage can be found online\footnote{http://mucar3.de/globalsip2018-gdpf}.
For computation, we used a standard office computer with an Intel(R) Core i7 CPU.
Moreover, the experiments were recorded with our institute's autonomous car MuCAR-3 \cite{feeb:mule} with roof-mounted Velodyne HDL64-S2.
Furthermore, for the ground truth data we installed an INS-sensor into our ground-truth object.

The goal of our evaluation is to estimate the $ x $ and $ y $ position of our ground truth object without id-switches.
We compare our results with the well-known Labeled Multi-Bernoulli Filter (LMB) \cite{bena:LMB}, its variation the Generalized-LMB (GLMB) \cite{bena:GLMB} and a classical filtering approach with an underlying track management \cite{himmelsbach2012tracking} denoted as BuTd.
As the other filters need clustered detections, we cluster the point cloud in coherent objects with the methods of \cite{bena:burger_iv2018,bena:BeNaIV2018,bib:burger_itsc2018}.
Then, the x and y positions of the bounding box centers $ \bm{z}_i^x $ and $ \bm{z}_i^y $ are used as measurements for the different methods.
Here, the measurement assignment likelihood of \Cref{eq:data_assi} is the following car-model based cluster prior:
\begin{equation} \label{eq:car_prior}
  \pi_{z_i(t)=k} =	e^{\left( -\left( \frac{(\bm{x}_x (t-1) - \bm{z}_i^x (t))^2}{a}+\frac{(\bm{x}_y (t-1) - \bm{z}_i^y (t))^2}{b}  \right)\right)},
\end{equation}
where $ \bm{x}_x (t-1) $ and $ \bm{x}_y (t-1) $ are the x and y position of the kth-component's state of time step $ t-1 $ and $ a,b \in \mathbb{R}$ are factors.
The ddCRP depends on the relationship (intersections, distance) between the bounding-box measurements, which is realized through the signed distance to the nearest side of the bounding-box to the state. An example can be seen in \Cref{fig:bb_ellipse,fig:bb}.

Furthermore, a simple constant velocity model with $ x,y $ positions and velocities as state variables for the dynamics of the tracks \cite{bena:BarShalom02} is applied.
At last, the tracked object which is the nearest to the ground truth object is evaluated in every frame.

Additional to the object based filtering, we implemented a grid based variant, which has grid cells as measurements.
In this case, the link assignment between neighboring grid cells for the ddCRP is utilized.
The measurement assignment likelihood is again the car-model based prior of \Cref{eq:car_prior}.
\Cref{fig:grid,fig:grid_ellipse} show an exemplary application of the filter variant.

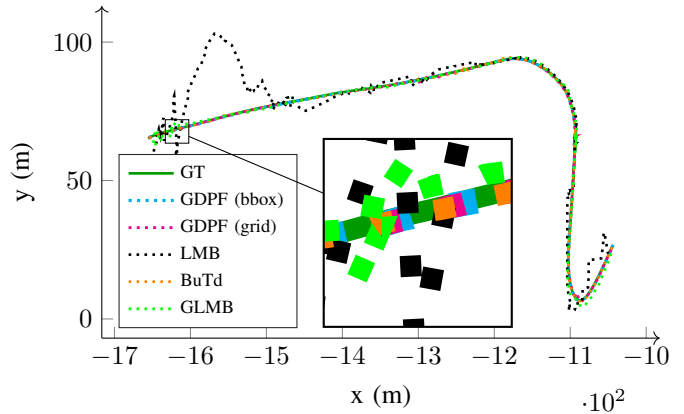
\begin{figure}[t!]
	\begin{center}
		\begin{tikzpicture}[spy using outlines={rectangle, magnification=8, size=1cm, connect spies}]
	\begin{axis}[
	height=6.cm,
	width=9cm,
	legend style={legend cell align=left, align=left, draw=white!15!black},
	legend pos = south west,
	axis x line*=bottom,
	axis y line*=left,
	scaled x ticks=base 10:-2,
	axis line style={->},
	xlabel=x (m),
	ylabel=y (m),
	]

	\addplot[color=green!60!black, line width=1pt] table[y=y,x=x,col sep=comma]{ground_truth_matrix.csv};
	\addplot[color=cyan,line width=1.2pt,dotted] table[y=y,x=x,col sep=comma]{dir_filter_tracked_objects.csv};
	\addplot[color=magenta,line width=1.1pt,dotted] table[y=y,x=x,col sep=comma]{dir_filter_grid_tracked_objects.csv};
	\addplot[color=black,line width=1pt,dotted] table[y=y,x=x,col sep=comma]{lmb_estimates_all_dets.csv};
	\addplot[color=orange,line width=1pt,dotted] table[y=y,x=x,col sep=comma]{multi_tracked_objects.csv};
	\addplot[color=green,line width=1pt,dotted] table[y=y,x=x,col sep=comma]{global_glms_5m.csv};
	\legend{\scriptsize GT,\scriptsize GDPF (bbox),\scriptsize GDPF (grid), \scriptsize LMB, \scriptsize BuTd, \scriptsize GLMB }

	\begin{scope}
		\spy[black,size=2.5cm] on (1.,2.75) in node [fill=none] at (4.2,1.4);
	\end{scope}

	\end{axis}
\end{tikzpicture}
	\caption{The tracked object's trajectories for every filter. The LMB particularly has problems at the initialization phase (black). The green dots correspond to the ground truth trajectory.} \label{fig:res_pos}
    \end{center}
\end{figure}

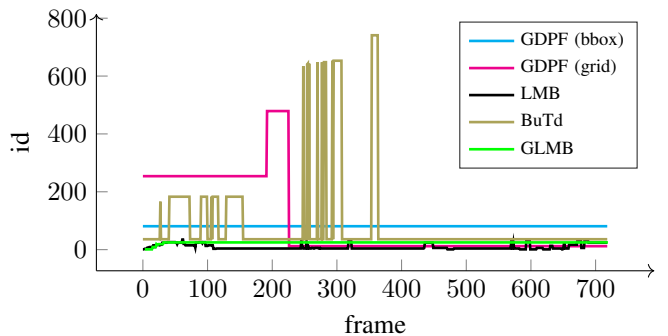
\begin{figure}[t!]
	\vspace{-0.5cm}
	\begin{center}
		\begin{tikzpicture}[mark size=1pt]
\begin{axis}[
height=5.cm,
width=9cm,
legend style={legend cell align=left, align=left, draw=white!15!black},
legend pos = north east,
axis x line*=bottom,
axis y line*=left,
axis line style={->},
xlabel=frame,
ylabel=id]
\addplot[color=cyan,line width=1pt] table[y=id,x=frame,col sep=comma]{dir_filter_tracked_objects.csv};
\addplot[color=magenta,line width=1pt] table[y=id,x=frame,col sep=comma]{dir_filter_grid_tracked_objects.csv};
\addplot[color=black,line width=1pt] table[y=id,x=frame,col sep=comma]{lmb_estimates_all_dets.csv};
\addplot[color=yellow!60!black,line width=1pt] table[y=id,x=frame,col sep=comma]{multi_tracked_objects.csv};
\addplot[color=green,line width=1pt] table[y=id,x=frame,col sep=comma]{global_glms_5m.csv};
\legend{\scriptsize GDPF (bbox),\scriptsize GDPF (grid), \scriptsize LMB, \scriptsize BuTd, \scriptsize GLMB}
\end{axis}
\end{tikzpicture}
	\caption{The tracked object's id for every filter. The optimal outcome is no id-switch at all, which means continuous tracking. The GDPF variants show good performances with only two and zero id-switches. In contrast, the BuTd and LMB approaches have many tracking fractions through a lot of id-switches.} \label{fig:ids}
	\end{center}
\end{figure}

\Cref{fig:res_pos,fig:ids} show that GDPF, in both variants, outperforms the other approaches. It smoothly follows the ground truth trajectory and has zero, respectively, two id-switches.
Contrary, the LMB and the GLMB have difficulties to stabilize the track at the beginning of the trajectory with resulting id-switches.
Moreover, the BuTd approach has a similar estimation quality of the position but many id-switches, which indicate track losses.
\Cref{tab:rmse} confirms the previous conclusions.
Our approach has the smallest RMSE, where BuTd has a similar result.
Furthermore, the LMB has a significant higher RMSE through the difficulties at the beginning of the test drive and $ 69 $ id-switches.
Moreover, we tested the LMB and GLMB with less detections, considering only detections which are in a $ 5 $ meter radius around the ground truth object.
Here, the LMB has a comparable result to the BuTd approach with an RMSE of $ 3.539 $ meter and $ 33 $ id-switches.
Furthermore, the GLMB approach has less id-switches but a higher RMSE.
Additionally, the mean computation time for all steps of the GDPF has been $ 34 $ms and $ 58 $ms (grid), for an average of $ 193 $ objects.

\begin{table}[t!]
	\begin{center}
	\begin{tabular}{c|c|c}
		Filter & RMSE & id-switches \\
		\hline
		GDPF (bbox) & 0.63657& 0\\
		GDPF (grid) & 0.89681& 2\\
		LMB  & 62.729& 69 \\
		LMB (low det)  & 3.539 & 33 \\
		GLMB (low det)  & 2.334 & 5 \\
		BuTd & 0.68749 & 31
	\end{tabular}
	\caption{RMSEs resulting from the position errors of the tracked objects and the corresponding id-switches throughout the test-drive.} \label{tab:rmse}
		\end{center}
	\vspace{-0.8cm}
\end{table}
\section{conclusion} \label{sec:conclusion}
In this paper, we propose the GDPF a multi-target tracker based on temporal dependent Dirichlet Process Mixture Models.
The GDPF has the ability to track an unknown number of targets in real-time with a probabilistic data association approach.
We demonstrated improved tracking results compared to other multi-target tracker.
Furthermore, we have shown the ability to handle non-clustered data, e.g, grid-cells as measurements as well as previously clustered objects.

Future work will focus on a classifying extension to utilize class specific priors in the association step.
Furthermore, we will investigate to integrate an extended object tracking by modeling the object appearance in the tracking part.
\bibliographystyle{IEEEtran}
\bibliography{IEEEabrv,additional_abrv,et_al,tas_papers,bena}

\begin{thebibliography}{10}
\providecommand{\url}[1]{#1}
\csname url@samestyle\endcsname
\providecommand{\newblock}{\relax}
\providecommand{\bibinfo}[2]{#2}
\providecommand{\BIBentrySTDinterwordspacing}{\spaceskip=0pt\relax}
\providecommand{\BIBentryALTinterwordstretchfactor}{4}
\providecommand{\BIBentryALTinterwordspacing}{\spaceskip=\fontdimen2\font plus
\BIBentryALTinterwordstretchfactor\fontdimen3\font minus
  \fontdimen4\font\relax}
\providecommand{\BIBforeignlanguage}[2]{{%
\expandafter\ifx\csname l@#1\endcsname\relax
\typeout{** WARNING: IEEEtran.bst: No hyphenation pattern has been}%
\typeout{** loaded for the language `#1'. Using the pattern for}%
\typeout{** the default language instead.}%
\else
\language=\csname l@#1\endcsname
\fi
#2}}
\providecommand{\BIBdecl}{\relax}
\BIBdecl

\bibitem{bena:BarShalom02}
Y.~Bar-Shalom, T.~Kirubarajan, and X.-R. Li, \emph{{Estimation with
  Applications to Tracking and Navigation}}.\hskip 1em plus 0.5em minus
  0.4em\relax New York, NY, USA: John Wiley \& Sons, Inc., 2002.

\bibitem{bena:mahlerstatistics}
R.~P. Mahler, ``{ "Statistics 101" for Multisensor, Multitarget Data Fusion},''
  \emph{{IEEE} Aerosp. Electron. Syst. Mag.}, vol.~19, no.~1, 2004.

\bibitem{BayesianFiltering}
S.~S{\"a}rkk{\"a}, \emph{{Bayesian Filtering and Smoothing}}.\hskip 1em plus
  0.5em minus 0.4em\relax Cambridge University Press, 2013, vol.~3.

\bibitem{bena:LMB}
S.~Reuter, B.-T. Vo, B.-N. Vo, and K.~Dietmayer, ``{The Labeled Multi-Bernoulli
  Filter},'' \emph{{IEEE} Trans. Signal Process.}, vol.~62, no.~12, Jun 2014.

\bibitem{bena:bleidistance}
D.~M. Blei and P.~I. Frazier, ``{Distance Dependent Chinese Restaurant
  Processes},'' \emph{Journal of Machine Learning Research}, vol.~12, 2011.

\bibitem{Wang_fastbayesian}
L.~Wang, D.~B. Dunson, B.~Branch, and M.~A, ``{Fast Bayesian Inference in
  Dirichlet Process Mixture Models},'' \emph{Journal of Computational and
  Graphical Statistics}, vol.~20, no.~1, 2011.

\bibitem{velodyne-lidar}
\BIBentryALTinterwordspacing
{Velodyne Lidar, Inc.}, ``{High Definition Lidar HDL-64E S2 Specifications}.''
  [Online]. Available:
  \url{{http://velodynelidar.com/lidar/hdlproducts/hdl64e.aspx}}
\BIBentrySTDinterwordspacing

\bibitem{bena:fries2017itsc}
C.~Fries, P.~Burger \emph{et~al.}, ``{How MuCAR Won the Convoy Scenario at
  ELROB 2016},'' in \emph{{Proc. IEEE Intelligent Transportation Syst. Conf.
  (ITSC)}}, Yokohama, Japan, Oct. 2017.

\bibitem{bena:burger_iv2018}
P.~Burger and H.-J. Wuensche, ``{Fast Multi-pass 3D Point Segmentation Based on
  a Structured Mesh Graph for Ground Vehicles},'' in \emph{{Proc. IEEE
  Intelligent Vehicles Symp. (IV)}}, Jun. 2018.

\bibitem{bena:BeNaIV2018}
B.~Naujoks and H.-J. Wuensche, ``{An Orientation Corrected Bounding Box Fit
  Based on the Convex Hull under Real Time Constraints},'' in \emph{{Proc. IEEE
  Intelligent Vehicles Symp. (IV)}}, 2018.

\bibitem{vo2006gaussian}
B.-N. Vo and W.-K. Ma, ``{The Gaussian Mixture Probability Hypothesis Density
  Filter},'' \emph{{IEEE} Trans. Signal Process.}, vol.~54, no.~11, 2006.

\bibitem{bena:vocardinality}
B.-T. Vo, B.-N. Vo, and A.~Cantoni, ``{The Cardinality Balanced Multi-Target
  Multi-Bernoulli Filter and its Implementations},'' \emph{{IEEE} Trans. Signal
  Process.}, vol.~57, no.~2, 2009.

\bibitem{granstromextended}
K.~Granstrom, C.~Lundquist, and O.~Orguner, ``{Extended Target Tracking using a
  Gaussian-Mixture PHD Filter},'' \emph{{IEEE} Trans. Aerosp. Electron. Syst.},
  vol.~48, no.~4, 2012.

\bibitem{bena:lmbextended}
M.~Beard, S.~Reuter \emph{et~al.}, ``{Multiple Extended Target Tracking with
  Labeled Random Finite Sets},'' \emph{{IEEE} Trans. Signal Process.}, vol.~64,
  no.~7, 2016.

\bibitem{bena:tuncersequential}
M.~A.~{\c{C}}. Tuncer and D.~Schulz, ``{Sequential Distance Dependent Chinese
  Restaurant Processes for Motion Segmentation of 3D Lidar Data},'' in
  \emph{Information Fusion (FUSION), 2016 19th International Conference
  on}.\hskip 1em plus 0.5em minus 0.4em\relax IEEE, 2016.

\bibitem{bena:foxddsdlm}
E.~Fox, E.~B. Sudderth, M.~I. Jordan, and A.~S. Willsky, ``{Bayesian
  Nonparametric Inference of Switching Dynamic Linear Models},'' \emph{{IEEE}
  Trans. Signal Process.}, vol.~59, no.~4, 2011.

\bibitem{bena:ddmotionseg}
M.~A.~{\c{C}}. Tuncer and D.~Schulz, ``{Sequential Distance Dependent Chinese
  Restaurant Processes for Motion Segmentation of 3D Lidar Data},'' in
  \emph{Information Fusion (FUSION), 2016 19th International Conference
  on}.\hskip 1em plus 0.5em minus 0.4em\relax IEEE, 2016.

\bibitem{bena:MU}
B.~Naujoks, T.~Engler, M.~Michaelis, and H.-J. Wuensche, ``{Measurement
  Uncertainty and its Influence on Dynamic Object Tracking in Autonomous
  Driving},'' in \emph{8. VDI-Fachtagung "Messunsicherheit 2017"}, Erfurt,
  Germany, Nov. 2017.

\bibitem{feeb:mule}
F.~Ebert, D.~Fassbender, B.~Naujoks, and H.-J. Wuensche, ``{Robust Long-Range
  Teach-and-Repeat in Non-Urban Environments},'' in \emph{{Proc. IEEE
  Intelligent Transportation Syst. Conf. (ITSC)}}, Yokohama, Japan, 2017.

\bibitem{bena:GLMB}
B.-N. Vo, B.-T. Vo, and H.~G. Hoang, ``{An Efficient Implementation of the
  Generalized Labeled Multi-Bernoulli Filter},'' \emph{{IEEE} Trans. Signal
  Process.}, vol.~65, no.~8, 2017.

\bibitem{himmelsbach2012tracking}
M.~Himmelsbach and H.-J. Wuensche, ``{Tracking and Classification of Arbitrary
  Objects with Bottom-Up/Top-Down Detection},'' in \emph{{Proc. IEEE
  Intelligent Vehicles Symp. (IV)}}.\hskip 1em plus 0.5em minus 0.4em\relax
  IEEE, 2012.

\bibitem{bib:burger_itsc2018}
P.~Burger, B.~Naujoks, and H.-J. Wuensche, ``{Fast Dual-Decomposition based
  Mesh-Graph Clustering for Point Clouds},'' in \emph{{Proc. IEEE Intelligent
  Transportation Syst. Conf. (ITSC)}}, Nov. 2018, accepted for publication.

\end{thebibliography}
\end{document}